\pdfoutput=1

\documentclass[11pt]{article}

\usepackage[preprint]{acl}

\usepackage{times}
\usepackage{latexsym}

\usepackage[T1]{fontenc}

\usepackage[utf8]{inputenc}

\usepackage{microtype}

\usepackage{inconsolata}

\usepackage{graphicx}

\title{Have we unified image generation and understanding yet? An empirical study of GPT-4o's image generation ability}

\author{Ning Li, Jingran Zhang, Justin Cui \\
  University of California, Los Angeles\\
    Los Angeles, CA 90095\\
  \texttt{\{ningli23, zhangjingran, justincui\}@ucla.edu}}

\begin{document}
\maketitle
\begin{abstract}
\noindent OpenAI's multimodal GPT-4o has demonstrated remarkable capabilities in image generation and editing, yet its ability to achieve world knowledge-informed semantic synthesis—seamlessly integrating domain knowledge, contextual reasoning, and instruction adherence—remains unproven. In this study, we systematically evaluate these capabilities across three critical dimensions: (1) \textit{Global Instruction Adherence}, (2) \textit{Fine-Grained Editing Precision}, and (3) \textit{Post-Generation Reasoning}. While existing benchmarks highlight GPT-4o's strong capabilities in image generation and editing, our evaluation reveals GPT-4o's persistent limitations: the model frequently defaults to literal interpretations of instructions, inconsistently applies knowledge constraints, and struggles with conditional reasoning tasks. These findings challenge prevailing assumptions about GPT-4o's unified understanding and generation capabilities, exposing significant gaps in its dynamic knowledge integration. Our study calls for the development of more robust benchmarks and training strategies that go beyond surface-level alignment, emphasizing context-aware and reasoning-grounded multimodal generation.
\end{abstract}

\section{Introduction}



The rapid evolution of multimodal AI has led to models capable of generating high-quality images from text, editing visual content with precision, and reasoning across modalities \cite{zhang2023text, chen2024next, zhao2023survey}. Among these, OpenAI’s GPT-4o represents a major advance, offering unified vision-language capabilities within a single architecture \cite{team2024chameleon, jin2023unified}. While its performance on standard benchmarks demonstrates impressive image generation and editing \cite{nichol2021glide, ghosh2024geneval, huang2023t2i, yan2025gpt-imgeval}, less is known about its ability to apply contextual reasoning, world knowledge, and instruction adherence during visual generation \cite{fu2024commonsense, meng2024phybench, sim2024evaluating}. For instance, GPT-4o may misinterpret spatial relationships (e.g., confusing left and right), fail to enforce abstract constraints, or produce content that violates basic facts \cite{sim2024evaluating, yuksekgonul2022and}. These observations motivate a closer empirical examination of how well such models integrate semantic understanding into image generation.

While recent research has focused on benchmarking text-to-image models for photorealism\cite{saharia2022photorealistic}, stylistic consistency, or textual alignment \cite{wu2023human, xu2024imagereward, lin2024evaluating}, few studies rigorously evaluate how well these systems understand the world they aim to represent \cite{fu2024commonsense, hu2024equip}. For instance, GPT-4o can generate "a dog left of a cat," but its ability to reinterpret spatial terms under reversed global instructions (e.g., mapping "left" to "right") or enforce exclusion constraints (e.g., avoiding out-of-scope objects like "coffee") remains unexplored \cite{sim2024evaluating}. These gaps raise questions about whether such models can dynamically apply world knowledge beyond surface-level pattern recognition \cite{yuksekgonul2022and}.

We conduct a systematic study of GPT-4o’s image generation through three lenses:  
\begin{enumerate}
    \item \textbf{Global Instruction Adherence}: Can the model override literal interpretations (e.g., spatial reversals)?  
    \item \textbf{Fine-Grained Editing Precision}: Does it modify elements while preserving context?  
    \item \textbf{Post-Generation Reasoning}: Can it retain and reason over visual context post-generation?
\end{enumerate}

Our experiments reveal persistent gaps: GPT-4o defaults to literal interpretations (e.g., ignoring reversed directions), inconsistently applies knowledge constraints, and struggles with contextual retention in conditional tasks. These findings challenge assumptions about unified understanding in multimodal LLMs and highlight the need for benchmarks that emphasize knowledge-guided synthesis and contextual generalization.  

\section{Related Work}
\textbf{Text-to-Image (T2I) Models}
Text-to-image (T2I) generation has advanced through specialized models prioritizing visual fidelity and multimodal architectures unifying vision-language capabilities. Specialized approaches evolved from autoregressive frameworks~\cite{chen2020generative,fan2024fluid} to diffusion-based methods~\cite{ho2020denoising,rombach2022high}, where latent diffusion models (LDMs) like Stable Diffusion~\cite{podell2023sdxl} balance efficiency and quality through iterative denoising in compressed latent spaces. Recent unified multimodal systems~\cite{team2024chameleon,jin2023unified,ge2024seed} extend large language models (LLMs) to visual generation via diffusion heads or autoregressive decoders, with architectures like Transfusion~\cite{zhou2024transfusion} demonstrating bidirectional diffusion-language integration for contextual alignment. Despite progress, both paradigms face challenges: specialized models lack dynamic reasoning beyond fixed text embeddings, while unified architectures struggle with inconsistent knowledge grounding~\cite{wu2024janus,ma2024janusflow}, limiting their semantic synthesis capabilities.  

\textbf{T2I Evaluations}
Evaluating text-to-image (T2I) models necessitates benchmarks that assess both technical fidelity and semantic alignment with real-world knowledge. Early efforts like GenEval~\cite{ghosh2024geneval} established foundational metrics for basic instruction compliance and photorealism, while Reason-Edit~\cite{huang2024smartedit} introduced instruction-guided editing tasks to evaluate fine-grained manipulation capabilities, albeit prioritizing syntactic edits (e.g., object replacement) over semantic reasoning. The WISE benchmark~\cite{niu2025wiseworldknowledgeinformedsemantic} addresses these limitations by pioneering world knowledge-informed evaluation, introducing 1,000 structured prompts across 25 subdomains (e.g., cultural common sense, natural science) and the WiScore metric to assess knowledge-image alignment beyond superficial text-pixel matching. While traditional benchmarks focus on image realism or object placement, WISE reveals significant deficiencies in models' ability to integrate domain knowledge, demonstrating that even unified multimodal architectures underperform dedicated T2I models in contextual accuracy—frequently generating anachronisms or scientific inaccuracies despite strong textual understanding capabilities. Similarly, \citet{zhang2025tweet} suggests to use conditional generation to evaluate the understanding of GPT-4o's generation ability, our work includes more types of prompts and evaluates GPT-4o more systematically.

\textbf{GPT-ImgEval}. The GPT-ImgEval benchmark~\cite{yan2025gpt-imgeval} presents the first comprehensive evaluation of GPT-4o’s image generation capabilities across three key tasks: text-to-image generation (via GenEval), instruction-based editing (via Reason-Edit), and world knowledge-informed synthesis (via WISE). GPT-4o is reported to demonstrate strong performance across all tasks, surpassing prior models in generation control, output quality, and knowledge reasoning. In addition to its quantitative findings, GPT-ImgEval conducts a classifier-based architectural analysis and offers empirical evidence suggesting GPT-4o uses a diffusion-based decoding head. The benchmark also notes generation-related limitations, including occasional content preservation issues, proportion control challenges, and synthesis artifacts, which offer insight into GPT-4o’s design and areas for improvement. Our works addresses GPT-4o’s capacity for semantic understanding instead, examining how well it integrates world knowledge, follows contextual instructions, and reasons across image generation tasks.

\begin{figure}
    \centering
    \includegraphics[height=11cm, keepaspectratio]{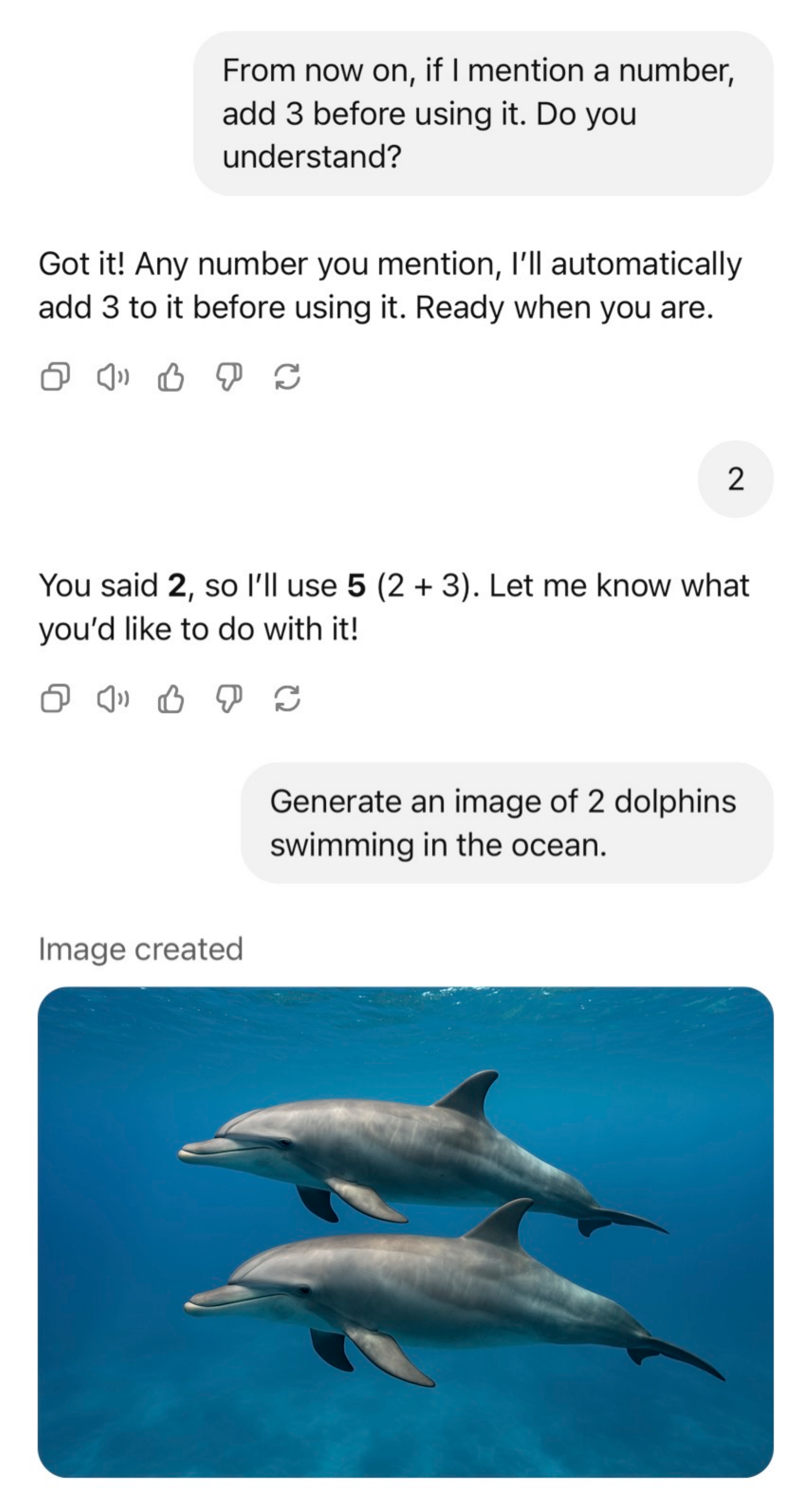}
    \caption{Demonstration of a global instruction prompt example.}
    \label{fig:dialogue}
\end{figure}

\section{Experiment}

\subsection{Prompt Construction}

To evaluate GPT-4o's image generation capabilities, we designed experiments based on three distinct types of prompts:

\textbf{Global Instruction.} In this prompt type, GPT-4o is provided with an overarching instruction prior to the image generation task. These global instructions introduce impactful contextual information that influences how subsequent prompts are interpreted. For instance, GPT-4o may be instructed to follow reversed spatial directions—such that "left" should be interpreted as "right" and vice versa. An example would be: although the prompt specifies “generate a cat on the left side of the image,” the model is expected to interpret this as “generate a cat on the right side of the image.” This type of prompt is designed to test whether the model can follow contextual logic when generating images, rather than relying solely on surface-level or literal interpretations of instructions.

\textbf{Image Editing.} As highlighted in previous work \citet{yan2025gpt-imgeval, huang2024smartedit}, the ability to edit images is a crucial aspect of evaluating a model's image generation performance. In this prompt type, GPT-4o is tasked with making specific edits to a given image—for example, removing or replacing certain elements. These prompts are designed to assess the model's capacity for fine-grained image manipulation, contextual consistency, and adherence to detailed visual instructions.

\textbf{Post-Generation Reasoning.} This type of prompt is designed to evaluate whether GPT-4o's reasoning ability is affected after transitioning into image generation mode. To assess this, the model is first prompted to generate a specific image. It is then given an additional prompt that requires either editing the previously generated image or producing a new image based on logical inference drawn from the initial output. For example, GPT-4o may first be instructed to generate an image of a zebra drinking water from a river. Subsequently, it is asked to generate an image of a man running on a road—but only if water is present in the previously generated image. This setup enables us to examine the model’s ability to retain contextual understanding and perform conditional reasoning after visual content has been produced.

\begin{figure*}[h!]
    \centering
    \includegraphics[width=\linewidth]{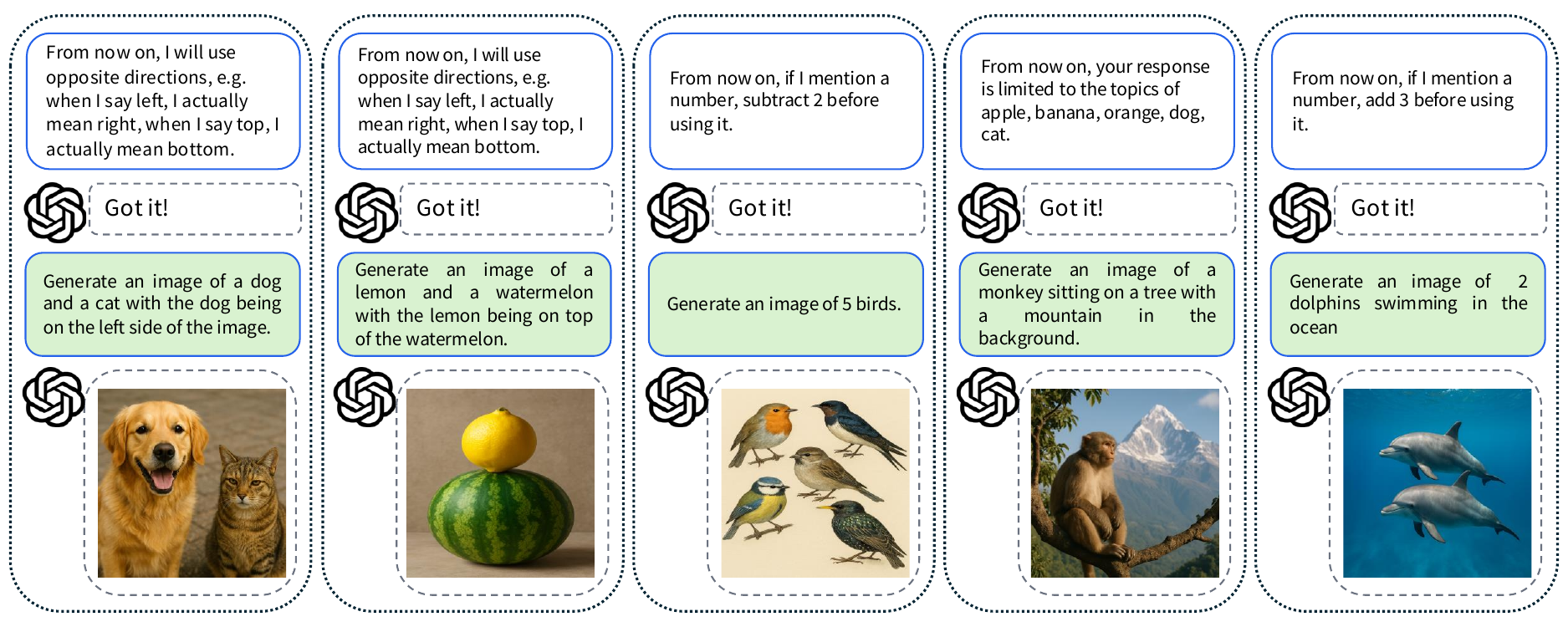}
    \caption{Examples of generated images with global instructions.}
    \label{fig:type1}
\end{figure*}

\subsection{Experimental Results}
\paragraph{Global Instruction Following}In Figure \ref{fig:type1}, we have demonstrated some examples of generated images with overarching global instructions. In the first two examples, where the instruction specifies reversed spatial directions (e.g., “left” means “right”), the model fails to comply, generating images with the dog correctly positioned on the literal left side. In the third and fifth examples, involving numerical transformations (subtracting or adding a value to mentioned numbers), GPT-4o again falls short—producing images with the exact number of objects (e.g., 5 birds and 2 dolphins) instead of the adjusted quantities. This suggests that while the model understands and executes literal prompts effectively, it struggles to integrate abstract numerical instructions. These results suggest that GPT-4o's image generation behavior is largely literal and often overlooks abstract global rules. A detailed interaction between us and GPT-4o can be found in Figure \ref{fig:dialogue}.

\begin{figure*}[h!]
    \centering
    \includegraphics[width=\linewidth]{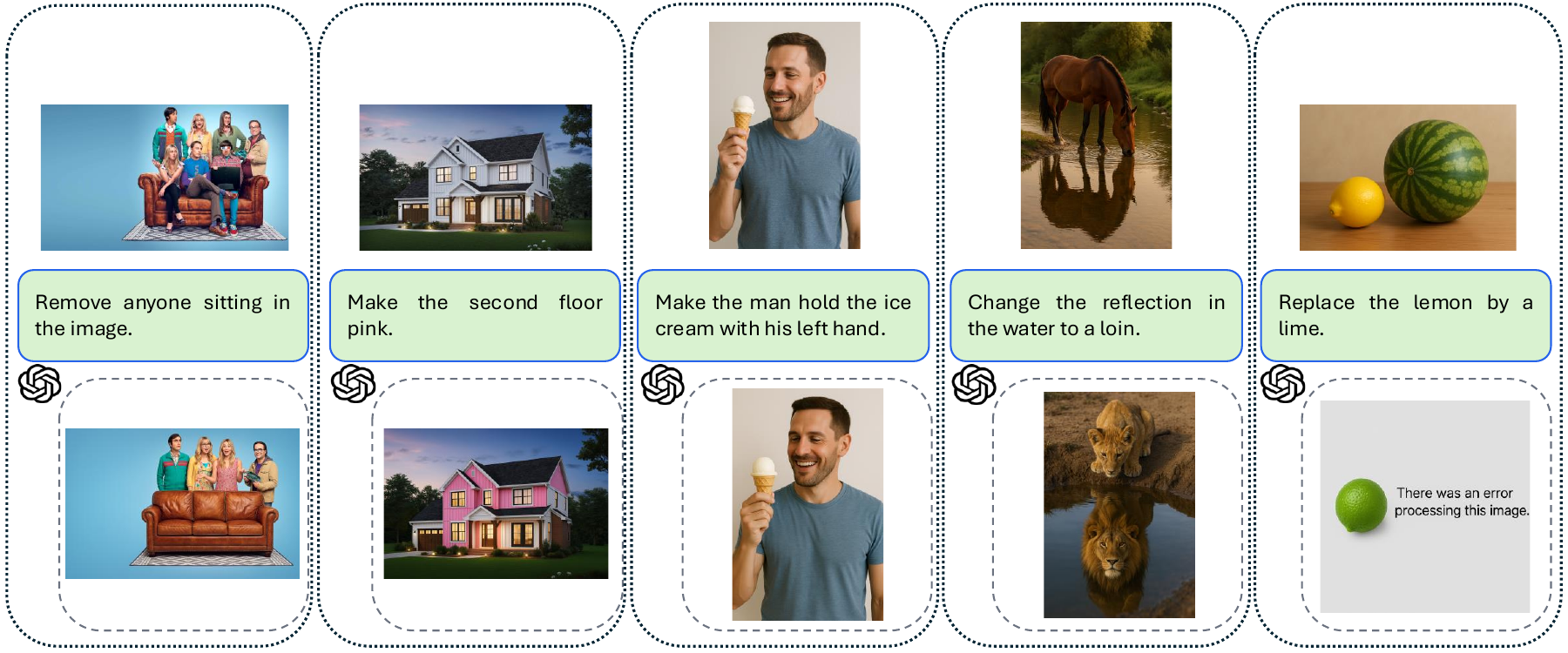}
    \caption{Examples of image editing performed by GPT-4o.}
    \label{fig:type2}
\end{figure*}

\paragraph{Image Editing Following} In Figure~\ref{fig:type2}, some examples of image editing performed by GPT-4o are presented. These examples reveal that while GPT-4o demonstrates a degree of capability in image manipulation, it often fails to fully comply with the given editing instructions. For instance, in response to the prompt requesting the removal of people sitting on a couch, the model not only removes the seated individuals but also mistakenly alters the people standing behind the couch. In another case, where the prompt asks to change the reflection of a horse in water to that of a lion, GPT-4o modifies both the horse and its reflection, rather than isolating the transformation to the reflection alone. These results suggest that GPT-4o may struggle with localized and selective image editing tasks, particularly when fine-grained spatial or semantic distinctions are required.

\begin{figure*}[h!]
    \centering
    \includegraphics[width=\linewidth]{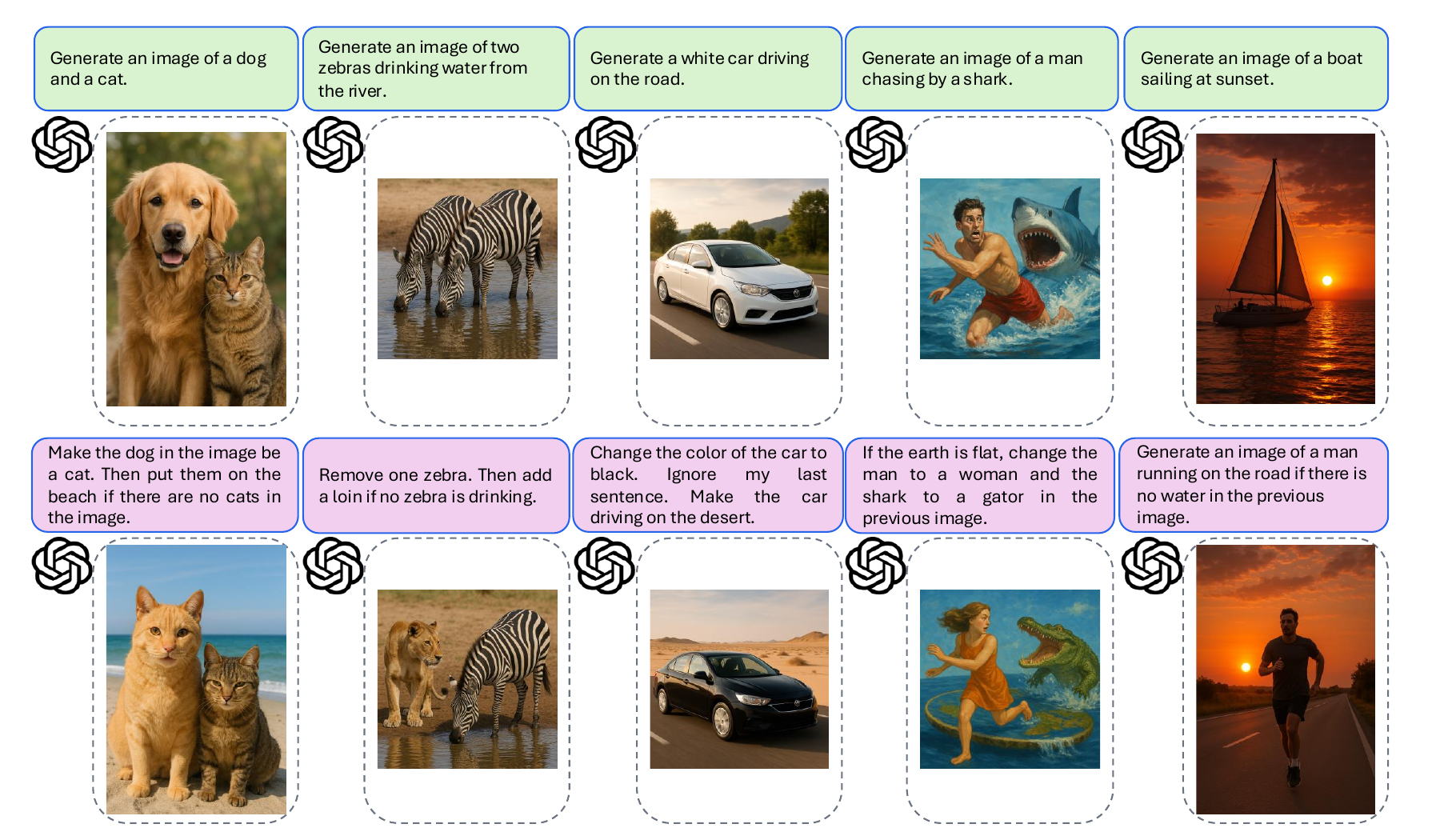}
    \caption{Examples of post-generation reasoning performed by GPT-4o.}
    \label{fig:type3}
\end{figure*}

\paragraph{Post-Generation Following}In Figure~\ref{fig:type3}, we present examples that reveal potential limitations in GPT-4o’s reasoning ability during image generation. In one scenario, the model is initially prompted to generate an image of a dog and a cat. In the follow-up prompt, it is instructed to replace the dog with a cat and move the resulting image to a beach setting—but only if there are no cats present in the original image. Since the original image already contains a cat, replacing the dog with another cat results in two cats, and thus, the condition for changing the background is not met. However, GPT-4o still performs both operations: it replaces the dog with a cat and also changes the background to a beach. This suggests that the model may struggle with interpreting and following conditional instructions that involve multi-step logical reasoning. Such errors indicate a limitation in GPT-4o’s ability to maintain logical consistency across sequential prompts within the image generation context.

\section{Conclusion}
Our empirical analysis reveals that GPT-4o, while capable of generating high-quality images and performing basic editing tasks, has not yet achieved true unification of image generation and understanding. Across three critical dimensions—global instruction adherence, fine-grained editing precision, and post-generation reasoning—the model frequently defaults to literal interpretations, overlooks abstract or contextual logic, and struggles with conditional reasoning. Our findings underscore the need for more robust benchmarks and training strategies that emphasize reasoning-aware generation, moving beyond surface-level alignment to foster deeper, context-sensitive multimodal intelligence.

\section{Future Work}
This study represents an initial step toward understanding the limitations of unified image generation and reasoning in multimodal models. In future work, we aim to include more types of prompts with broader coverage across diverse reasoning and generation scenarios. Furthermore, our analysis will extend beyond GPT-4o to include a range of state-of-the-art multimodal models. By benchmarking across multiple systems, we seek to identify common failure modes and better characterize the broader challenges in achieving true unification of image generation and reasoning.

\bibliography{custom}




\end{document}